\documentclass[oribibl]{llncs}

\usepackage{multicol}
\usepackage{amsmath}
\usepackage{tabu}
\usepackage{amssymb}
\usepackage{graphicx}
\usepackage{epsfig}
\usepackage{hhline}
\usepackage{times}
\usepackage{graphicx}
\usepackage{fancybox}
\usepackage{color}
\usepackage{multicol}
\usepackage{xspace,epic,eepic}
\usepackage{latexsym}

\newcommand{\ignore}[1]{}

\newcommand{\ul}[1]{\underline{#1}}
\newcommand{\boxtheorem}{\hfill $\Box$}
\newcommand{\nit}[1]{{\it #1}}

\newcommand{\mc}[1]{\mathcal{ #1}}

\newcommand{\mbb}[1]{\mathbb{ #1}}

\newcommand{\msf}[1]{\mathsf{ #1}}

\newcommand{\e}{\mathbf{e}}
\newcommand{\C}[1]{\mathcal{C}}
\newcommand{\T}[1]{\mathcal{T}}

\newcommand{\Resp}{\nit{Resp}}
\newcommand{\Shap}{\nit{Shap}}

\newcommand{\red}[1]{\textcolor{red}{#1}}
\newcommand{\re}[1]{\textcolor{red}{#1}}

\newcommand{\bl}[1]{\textcolor{blue}{#1}}

\newcommand{\comlb}[1]{{\vspace{2mm}\noindent \bf  {\red{COMM(LEO):}}}~ #1 \hfill {\bf
    END.}\\}

\title{Attribution-Scores and Causal Counterfactuals as Explanations in  Artificial Intelligence}

\author{{\bf  Leopoldo \ Bertossi}\thanks{{\bf leopoldo.bertossi@skema.edu}. \ Member of the Millennium Institute for Foundations of Data Research (IMFD, Chile).}}

\institute{{\bf SKEMA Business School}\\{\bf Montreal, Canada}}

\begin{document}

\thispagestyle{empty}
\pagestyle{plain}
\maketitle

\begin{abstract}
In this expository article we highlight the relevance of explanations for artificial intelligence, in general, and for the newer developments in {\em explainable AI}, referring to origins and connections of and among different approaches. We describe in simple terms, explanations in data management and machine learning that are based on attribution-scores, and counterfactuals as found in the area of causality. We elaborate on the importance of logical reasoning when dealing with counterfactuals, and their use for score computation.
\end{abstract}

\section{Introduction}\label{sec:intro}

The search for explanations  belongs to human nature, and as a quest, it has been around since the inception of human beings. Explanations, as a subject,  have been investigated in Artificial Intelligence (AI) for some decades, and also, and for a much longer time, in other disciplines, such as Philosophy, Logic, Physics, Statistics. Actually, the explicit study of explanations can be traced back to the ancient Greeks, who were already concerned with {\em causes} and {\em effects}.
\ Nowadays, a whole area of artificial intelligence (AI) has emerged, that of {\em Explainable AI} (XAI). It has become part of an even larger area, {\em Ethical AI}, that encompasses other concerns, such as  fairness, responsibility, trust, bias (better, lack thereof), etc.

Research done under XAI has become diverse, intensive, extensive, and definitely, effervescent. Accordingly, it is difficult to keep track of: (a) the origins of some popular  approaches to explainability; and (b) the new methodologies, and the possible connections, similarities and differences among them. \ It is also the case that AI has become of interest for many stakeholders, and most prominently, for people who may be affected by developments in AI and by the use of AI-based systems. It has also become a subject of investigation or discussion for people who do not directly do  research,  system development or applications inside AI. This has contributed to a certain degree of confusion about what exactly falls under AI, and the role of XAI within AI.

Consistently with these observations, in this work we will revisit, in intuitive and simple terms, some ``classic" approaches to explanations that have been introduced and investigated in AI, and have been established  in it for some time. We will also discuss newer approaches to explanations that have emerged mainly in the context of Machine Learning (ML), and have become best examples of  Explainable AI. \
One of our goals  is to make clear that explanations  have been investigated in AI much before this ``new wave" of XAI arrived \cite{molnar}, and that the new methods build (or can be seen as building) upon older approaches. \ We also discuss in more detail the place of XAI in AI at large.

Motivated by our own recent research, we describe some approaches to XAI that are based on assigning numerical scores to elements of an input to an AI system, e.g. an ML-classifier,  that give an account of their relevance with respect to the outcome obtained from the system for that particular input. Instead of providing all the mathematical and algorithmic background, we use concrete examples to convey the main ideas and issues. Technical details can be found in the provided bibliographic references.

We end this article with a general discussion of the role of reasoning in XAI as it has been traditionally understood in AI. In general, the current approaches do not explicitly appeal to reasoning, nor are they extended with any kind of logical reasoning. We argue in favor of extending these newer approaches with reasoning capabilities, in particular, with counterfactual reasoning, which has to do with exploring, analyzing and comparing, usually hypothetically,  different alternative scenarios. This form of reasoning is at the basis of causal explanations, and other areas of computer science.

 This is not an exhaustive survey of explanation methods, nor of XAI. Recent surveys that cover XAI can be found, e.g. in \cite{Burkart,fosca,minh,molnar}. This expository article can be taken as an invitation and a basis for a broader discussion around explanations, and explainable AI; and also as a motivation to explore some subjects in more detail.


\section{The Role of Explanations in AI}\label{sec:role}

Explanations are an important part of Artificial Intelligence (AI), and we can clearly identify a couple of fundamental reasons for this:
\begin{itemize} \item[(A)] Searching for explanations for external phenomena, observed behaviors, etc., and providing them , are important manifestations of human intelligence, and as such, they become natural subjects of investigation in AI.
\item[(B)] AI systems themselves provide results of different kinds, and they require explanations, for humans and AI systems as well.
\end{itemize}
 The former direction has been investigated in AI and more traditional disciplines, as already mentioned above.  The second direction is much newer, because AI systems used at large are also much newer. The ``older" kind of explanations can be used and adapted for this second purpose, but also some new forms of explanations, that can be more {\em ad hoc} for different kinds of AI systems, have been introduced and investigated in the last few years. Let us consider a simple example, for the gist.

\begin{example} \label{ex:one} Consider a client of bank who is applying for loan. The bank will process his/her application by means of an AI system that will decide if the client should be given the loan or not. In order for the application to be computationally processed, the client is represented as an {\em entity}  describing him/her, say \ $\e = \langle \msf{john}, \msf{18}, \msf{plumber}, \msf{70K},$ $ \msf{harlem}, \msf{10K}, \msf{basic}\rangle$, that is, a finite record
of  {\em feature values}. The set of features is $\mc{F} = \{\msf{Name}$, $\msf{Age}$, $\msf{Activity}$, $\msf{Income}$, $\msf{Debt}$, $\msf{EdLevel}\}$.

The bank uses a {\em classifier}, $\mc{C}$, that is an AI system that may have been learned on the basis of existing data about loan applications. There are different ways to build such a system. After receiving input $\e$, \ $\mc{C}$ returns a {\em label}, {\em Yes} or {\em No} (or $0$ or $1$). In this case, it returns {\em No}, indicating that the loan request is rejected. \ The client (or the bank executive) asks {\em ``Why?"}, and would like to  have an explanation. The issues are: (a) What kind of explanation? \ (b) How could it be provided? \ (c)
From what? \boxtheorem
\end{example}

These kinds of motivations and applications are typical of {\em Explainable AI} (XAI)  these days, and of explainable machine learning, in particular. Actually, the whole area has become part of a larger one, usually called {\em Ethical AI}, which includes concerns such as transparency, fairness, bias, trust, responsibility, etc., that should be taken into account when the use of AI systems may affect  stakeholders. I's no wonder that these issues are being discussed and investigated in other disciplines, such as
 Law, Sociology, Philosophy; and others that are more directly affected by the use of AI systems, e.g. Business, Medicine, Health, etc. Some countries have already passed new legislation forcing AI systems affecting users to provide guarantees of an ethical behaviour \cite{fosca2}. {\em Explainability  and interpretability} of AI systems  are  part of this picture.

 Some claim that ethical AI research is {\em not part of} AI, but {\em about} AI. The fact that an increasing number of people who work in this area do not do actual AI research, does not make the area less AI. It is part of AI for several reasons, among them: (a)
AI systems should be extended with the capability to provide explanations, and the extended systems would become also AI systems; (b) the individual subjects that fall under ethical AI are developed on a scientific and technical basis by AI researchers, who understand, model and implement explanations; (c) as already mentioned, explanation finding and giving are intelligent human activities worth of investigation under AI; (d) explanations become additional resources for AI system building, e.g. one can (automatically) learn from explanations; etc.

\section{Some Classical Models of Explanation}\label{sec:expl}

In this section, we will briefly introduce and discuss some approaches to explanations that have been proposed and investigated in the context of AI, actually for a few decades by now. Some of them fall in the area of AI called {\em model-based diagnosis} \cite{struss}, in that the explanation process relies on the use of a mathematical model of a system under observation, e.g. a logical or a probabilistic model. We will use a running example to illustrate different approaches.

\subsection{Consistency-based diagnosis}\label{sec:cbd}

 If we are confronting a system that is exhibiting an unexpected or abnormal behavior, we want to obtain a {\em diagnosis} for this, i.e. some sort of explanation. Diagnoses are obtained from a model of the system, possibly extended with some additional knowledge. In the following, we briefly describe the approach to diagnosis proposed by Ray Reiter \cite{reiterDiag}, usually called {\em consistency-based diagnosis} (CBD).

 \begin{example} \label{ex:circ} Consider the very simple Boolean circuit in Figure \ref{fig:circ}, with an {\em And}-gate, $A$, and an {\em Or}-gate, $O$. The input variables are $a,b,c$, the intermediate output variable for $A$ is $x$, and the final, output variable is $d$; all of them taking values $0$ or $1$. The intended meaning of the propositional variable $a$ is ``input $a$ is true" (or takes value $1$), etc.

\begin{figure}[h]
 \centerline{\includegraphics[width=5.5cm]{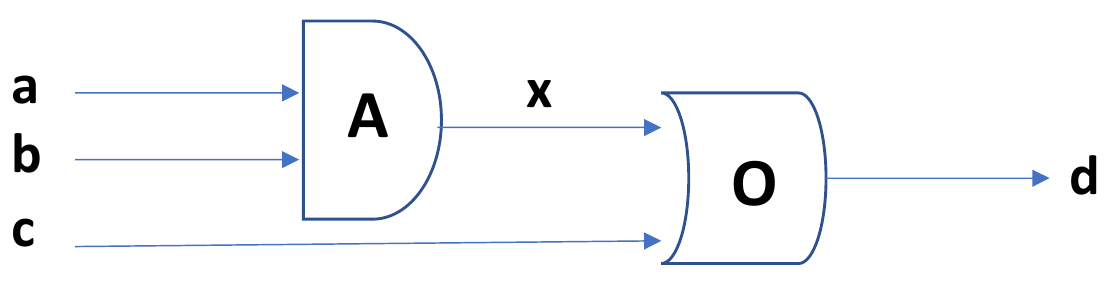}}
 \caption{A Boolean circuit}\label{fig:circ}
\end{figure}

From a conceptual point of view, this could be a {\em binary classifier}, for which an input entity $\e$ is represented by a record of binary values for $a,b,c$. The classifier computes the output $d$, which becomes the binary label assigned to $\e$. \
Now, assume we observe the following behavior:

\begin{figure}
 \centerline{\includegraphics[width=5.5cm]{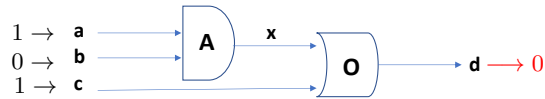}}

\vspace{-1.25cm}\noindent {\footnotesize \hspace*{2.5cm} $1 \rightarrow$\\ \vspace{-0.7mm}
\hspace*{2.5cm}$0 \rightarrow$\\ \vspace{-0.7mm}
\hspace*{2.53cm}$1 \rightarrow$}

\vspace{-0.6cm}\hspace*{8.8cm}  $\re{\longrightarrow 0}$

\vspace{5mm}
  \caption{A malfunctioning Boolean circuit}\label{fig:circMal}
\end{figure}

\vspace{-5mm}
According to the laws of propositional logic, we should obtain $x=0$ and $d=1$, which is not the case here. \ We ask: \ {\em Why?} \ {\em What is wrong?} \ Or more specifically, {\em What is a {\em diagnosis} for the abnormal behavior of the circuit?}
\boxtheorem
\end{example}

A more general  question now is: {\em What is a diagnosis?} \ Diagnoses have to be characterized. Since we are adopting a  model-based approach, we need a model. \ In our example,
a logical model of the circuit, when it works properly, is the set of propositional formulas
\begin{equation} \{(x \longleftrightarrow (a \wedge
b)),  \ (d \longleftrightarrow (x \vee
c))\}. \label{eq:cir}
\end{equation} However, our circuit at hand, by working abnormally, is {\em not} modeled by these formulas. \

Furthermore, notice that
the {\em observation}, $\nit{Obs}$, represented by the formula: \ $ a \wedge \neg b \wedge c \wedge \neg d$, indicating that $a$ and $c$ are true, $b$ is false, and $d$ is true, is mutually inconsistent with the ideal model above, that is, there is no assignment of truth values to the propositional variables that makes the combination true. From their combination we cannot logically obtain any useful information, because everything is entailed by an inconsistent set of propositional formulas. We may want instead {\em a model that allows failures, or abnormal behaviors}. From such a model, we could try to obtain explanations for those abnormal behaviors.

A better, more flexible model  that allows failures, and specifies how components behave under normal conditions is as follows:
\begin{equation}
\mc{M} = \{\neg AbA \longrightarrow (x \leftrightarrow (a \wedge
b)), \ \ \neg AbO \longrightarrow (d \leftrightarrow (x \vee
c))\}. \label{eq:model}
\end{equation}
The first formula says
{\em ``When $A$ is not abnormal,  it works as an
And-gate"}, etc. Here, $\nit{AbA}$ and $\nit{AbO}$ are new propositional variables.
This is a ``weak model of failure" in that it specifies how things behave under normal conditions only, but not under abnormal ones. This is a typical (but non-mandatory) form models take under CBD \cite{ray92}. The model assumes that the only potentially faulty components are, in this case, the gates (but not the wires connecting them), a modeling choice. \
\ Now gates could be abnormal (or faulty), and $\{\nit{Obs}\} \ \cup \ \mc{M}$ \ is a perfectly consistent logical model.

Notice that when one specifies, in addition, that the gates are not abnormal, we have:
\begin{equation}
  \{\nit{Obs}\} \ \cup \ \mc{M} \ \cup \{\neg \nit{AbA}, \neg \nit{AbO}\} \ \mbox{ is \ inconsistent}, \label{eq:inco}
  \end{equation}
as before. So, something has to be abnormal. It is through {\em consistency restoration} that we will be able to characterize and compute diagnoses.

Notice that making gate $\bl{O}$ abnormal in (\ref{eq:inco}) restores consistency, that is, in contrast to (\ref{eq:inco}),
 \begin{equation*}
\nit{Obs} \ \cup \ \mc{M} \ \cup \{\neg \nit{AbA}, \ \ul{\nit{AbO}}\} \ \mbox{ is \ consistent}.
\end{equation*}
We are underlying the change we made. More precisely,  and by definition, $\{\nit{abO}\}$ is a diagnosis. \ Similarly,
 $D^\prime = \{\nit{abO}, \nit{abA}\}$ is a diagnosis, because making every gate abnormal also restores consistency:

\begin{equation*}
\nit{Obs} \ \cup \ \mc{M} \ \cup \{\ul{\nit{AbA}}, \ \ul{\nit{AbO}}\} \ \mbox{ is \ consistent}.
\end{equation*}

We may consider $D$ as a ``better" diagnosis than $D^\prime$, because  it makes fewer assumptions, is more informative by providing narrower and more focused diagnosis. $D$ is a {\em minimal diagnosis} in that it is not set-theoretically included in any other diagnosis. It is also a {\em minimum diagnosis} in that it has a minimum cardinality. As expected, different preference criteria could be imposed on diagnosis.

\subsection{Abduction}

Abduction is a much older approach to obtaining a {\em best} explanation for an observation at hand. It can be traced back to Aristotle, and, more recently, to the work by the Philosopher C.S. Peirce \cite{peirce}.   Abductive diagnosis has been extensively investigated in AI, and has become one of the classic approaches to explanations \cite{pooleBook,struss,marquis}.

For the gist, and at the risk of simplifying things too much, a typical example provides the following simple model (a propositional logical theory):
\ $\nit{Covid19} \ \rightarrow \ \nit{Breathlessness}$. Now, we observe (for a patient): \ $\nit{Obs}\!: \ \nit{Breathlessness}$. In the light of the only information at hand, that provided by the model, we may ``infer" $\nit{Covid19}$ as an explanation. However, this is not classical inference in that we are using the implication backwards. So, $\nit{Covid19}$ is being {\em abduced} from the model and the information (as opposed to classically inferred or deduced), in the sense that:
 \begin{equation}
\{\ul{\nit{Covid19}}\} \ \cup \ \{\nit{Covid19} \ \rightarrow \ \nit{Breathlessness}\} \ \models \ \nit{Obs}, \label{eq:abd}
\end{equation} which defines $\nit{Covid19}$ as an abductive explanation.  \ Here, the symbol $\models$ denotes classical logical consequence.\footnote{If some other non-classical logic is used instead, $\models$ has to be replaced by the corresponding entailment criterion \cite{eiter}.} \ Of course, if the model becomes more complex,
 this sort of {\em backward reasoning} in search for explanations that support implications (via forward reasoning),
becomes much more complex and computationally costly. As expected, one may consider additional  preference criteria on abductive explanations, most typically some sort of minimality condition.

 \ignore{
 Abduce (not deduce) abnormalities that imply the observation: \ roughly

 \vspace{1mm}
 $\re{\nit{abO}} \ \cup  \ \mc{M} \ \cup \ \{ a \wedge \neg b \wedge c\} \ \cdots \cup \ \re{\rightarrow \ \neg d}$
 }

\begin{example} \ (example \ref{ex:circ} cont.) \ Consider the same circuit and observation, in this case $\nit{Obs}^\prime = \{\neg d\}$, which we would like to entail with additional information provided by abductible facts.  We cannot expect to obtain \ $\{\ul{\nit{abO}}\}  \cup   \mc{M} \ \cup \ \{ a \wedge \neg b \wedge c\} \ \models \ \neg d$, with $\mc{M}$ as in (\ref{eq:model}), which is only a weak model of failure. \ Actually, the entailment does not hold since $\{\nit{abO}\}  \cup   \mc{M} \ \cup \ \{ a \wedge \neg b \wedge c\} \ \cup \ \{d\}$ is a consistent theory.

\ignore{
In fact, from the model in (\ref{eq:model}), we obtain, in particular, $\nit{AbO} \rightarrow  (\neg d \leftrightarrow \neg (x \vee c))$.
\ Now, the RHS of the double implication is false (because $c$ is true). Then, we have $\nit{AbO} \rightarrow  (\neg d \leftrightarrow \nit{false})$. The only way in which $\neg d$ can be true (the observation), without making $\nit{AbO} \rightarrow  (\neg d \leftrightarrow \nit{false})$ false (which we assumed, as a part of the model, to be true) is that $\nit{AbO}$ becomes false. That is, for $\neg d$ to be true, $\neg \nit{AbO}$ has to be true.}

However, if we change the model in order to specify failures, e.g. with \ $\mc{M}^\prime = \{a \wedge b \wedge \nit{AbA} \rightarrow \neg x, \ \ x \wedge \nit{AbO} \rightarrow \neg d, \ \ c \wedge \nit{AbO} \rightarrow \neg d \}$, \
we do obtain \ $\{\ul{\nit{abO}}\} \cup \mc{M}^\prime \cup \{a,\neg b, c\} \models \neg d$,
as expected. It is common, but not mandatory, to use abduction with implicational models \cite{eiter2}.
\boxtheorem
 \end{example}
Abduction has found its way into XAI, most prominently, to provide explanations for results from classification models. They are usually called {\em sufficient explanations}: \ A (hopefully minimal) set of abducible facts that are sufficient to entail the observed label \cite{marques-silva19, marques-silva,barcelo}.

\ignore{++++++++++++++++++++++++++++++++XXXX+++++++
\comlb{Check below. Poole's paper on Abduction on probabilistic Horn theories. Y el paper de ASP en diagnosis de Eiter et al.}

We have considered so far models written in classical logic. As such they are not probabilistic models. In knowledge representation in general, a probabilistic model could specify probabilistic dependencies among variables \cite{russell,luc}. For example, among other kinds of models, a Bayesian network does this job by considering a directed graph of variables with some additional conditional and absolute probabilities for them, and some additional independence assumptions (that are compatible with the graph structure). More generally, with a probabilistic model, on which basis the variables have a joint distribution $P$ (explicit or not), with associated conditional distributions, abductive explanations are those (abducible) variables $V$ that that maximize the conditional probability, $P(\nit{Obs}|V)$, of the observation. However, its is more common with this kind of models to find those variables $V$ that are best explanations in the sense that for the {\em posterior probability}, $P(V|\nit{Obs})$, i.e. given the observation, is maximized.
+++++XXXXX++++}

\ignore{++++
There are  connections between these  forms of MBD
}

\subsection{Actual causality and responsibility}\label{sec:causes}

We can use the running example to illustrate the notion of {\em actual causality} \cite{HP05,halpern}.
 \begin{example} \ (example \ref{ex:circ} cont.) \label{ex:cause} \ Consider the same circuit as in Figure \ref{fig:circMal}, and the same model as in (\ref{eq:model}), for which we had in (\ref{eq:inco}):
$$\{a, \neg b, c\} \ \cup \ \mc{M} \ \cup \{\neg \nit{AbA}, \neg \nit{AbO}\} \ \cup \  \{\neg d\} \  \mbox{ is inconsistent},$$
which is logically equivalent to:
\begin{equation}
\{a, \neg b, c\} \ \cup \ \mc{M} \ \cup \{\neg \nit{AbA}, \neg \nit{AbO}\} \ \models \ d.  \label{eq:count}
\end{equation}
In this setting, we will play a {\em counterfactual} game consisting in hypothetically changing variables' truth values, to see if the entailment in (\ref{eq:count}) changes. Before  proceeding, we have to identify the {\em endogenous} variables, those on which we have some control, and the {\em exogenous variables} we have as a given.   This choice is application dependent. In our case, it is natural to consider $a, b,c$ as exogenous, and $x, d, \nit{abA}, \nit{abO}$ as endogenous. Here, the {\em interventions} are the hypothetical changes of non-abnormalities into abnormalities, to see if implication in (\ref{eq:count}) changes as a result.  (The  interventions are also application dependent.)

Switching $\neg \nit{abO}$  into $\nit{AbO}$, does invalidate the previous entailment:
\begin{equation*}
\{a, \neg b, c\} \ \cup \ \mc{M} \ \cup \{\neg \nit{AbA}, \ul{\nit{AbO}}\} \ \not \models \ d.  
\end{equation*}
For this reason (and by definition), we say that ``$\nit{abO}$  is a {\em counterfactual cause}" (for the observation).

  However, when we switch $\nit{abA}$: \ $\{a, \neg b, c\} \ \cup \ \mc{M} \ \cup \{\ul{\nit{AbA}},  \neg \nit{AbO}\} \    \models \ d$. The entailment still holds. Accordingly, $\bl{\nit{AbA}}$ is {\em not} a counterfactual cause. For this candidate, an extra counterfactual change, a so-called {\em contingent} change, is necessary:
\ $\{a, \neg b, c\} \ \cup \ \mc{M} \ \cup \{\ul{\nit{AbA}},  \ul{\nit{AbO}}\}  \ \not \models \ d.$ \ Had $\nit{abO}$ not been already (and alone) a counterfactual cause, $\nit{AbA}$ would have been called (by definition) an  {\em actual  cause with contingency set} $\{abO\}$. \ So, $\nit{AbA}$ is neither a counterfactual nor an actual cause.\footnote{Example \ref{ex:mbdaex5} will show an actual cause that is not a counterfactual cause.} \boxtheorem
 \end{example}

Actual causality provides {\em counterfactual explanations} to observations. In general terms, they are ``components" of a system that are a cause for an observed behavior. Counterfactual causes are actual causes with an empty contingency set. Accordingly, counterfactual causes are {\em strong} causes in that they, by themselves, explain the observation. Actual (non-counterfactual) causes are {\em weaker} causes, they require the company of other components to explain the observation.

Readers who are more familiar with causality based on {\em structural models} \cite{pearl,roy} may be missing them here. Actually, the diagnosis problems can be cast in those terms too. A purely logical model, as in the previous examples, does not distinguish causal directions, or between causes and effects. They can be better represented by a structural model that takes the form of a (directed) {\em causal network}.

\begin{example} (example \ref{ex:cause} cont.) The following causal network represents our diagnosis example, or better, our possibly faulty circuit.

\begin{figure}
 \centerline{\includegraphics[width=7.3cm]{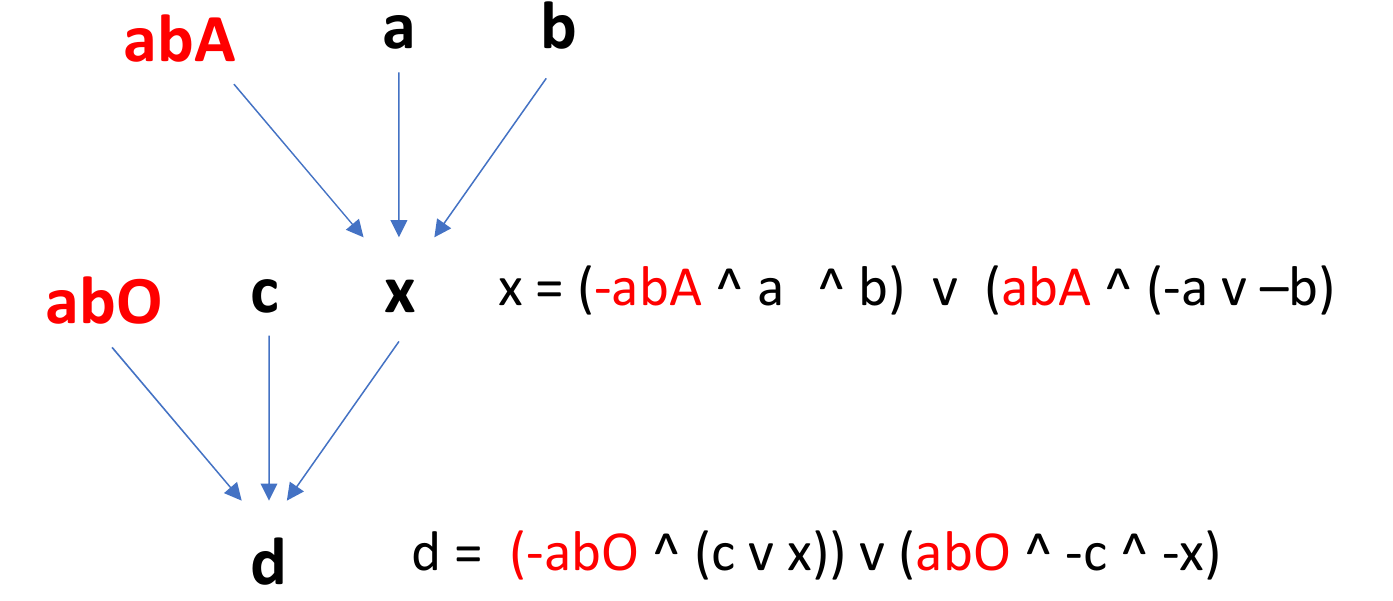}}
 \vspace{-2mm}\caption{Causal Network}\label{fig:net}
 \end{figure}

Here, $\nit{abA}, \nit{abO}$ are endogenous variables, which can be subject to counterfactual interventions, which, in this case, amount to making $\nit{AbA}$ and $\nit{AbO}$ true or false.
\ Variables $x$ and $d$ are endogenous, and have {\em structural equations} associated to them, as shown in Fig. \ref{fig:net}, capturing the circuit's logic. They are unidirectional, so as the edge directions. Contrary to the weak model of failure in (\ref{eq:model}), the equations specify the behaviour also under abnormality conditions. \boxtheorem
\end{example}

Actual causality has been extended with the quantitative notion of {\em responsibility} \cite{Chockler04}. Causes are assigned a numerical score that captures their causal strength. The score takes into account, for endogenous variables, if it is an actual cause or not; and, in the positive case, its contingency sets (CSs).

\begin{example} \ (example \ref{ex:cause} cont.) \ The responsibility score for an endogenous variable takes into account , in this case, is defined (for the endogenous variables) as follows: \vspace{-1mm}
 \begin{eqnarray*}
 \nit{Resp}(\nit{abO}) &:=& \frac{1}{1 + |\Gamma|} = \frac{1}{1 + 0} = 1,\\
\vspace{-2mm} && \hspace*{1mm}\mbox{ with $\Gamma$ a minimum-cardinality CS for } \nit{abO}.\\
\nit{Resp}(\nit{abA})  &~:= 0&.
 \end{eqnarray*}

\vspace{-3mm} The first case is due to the fact that $\nit{abO}$ is a counterfactual cause, and then, has the maximum responsibility, $1$. The second case is due to the fact that $\nit{abA}$ is not an actual cause.\footnote{Less ``trivial" cases will be shown in Example \ref{ex:mbdaex5}.}
 \boxtheorem
\end{example}

Actual causality has been applied to provide explanations in the context of relational databases \cite{suciu10,suciuDEBull,tocs}. There, the observation is a query result, and one wants to unveil the causes, inside the database, in the form of, say, database tuples or attribute values in them, for the query to be true (or returning a particular answer). Also, responsibility scores can be assigned to tuples or their attribute values.

\begin{example}\label{ex:mbdaex5} Consider the relational database instance $D$:

\begin{multicols}{2}
\begin{center}
$\begin{tabu}{|c|}
\hline R\\ \hline \langle c, b \rangle \\ \langle a, d \rangle \\ \langle b, a\rangle \\
\hline
\end{tabu} \hspace{5mm}\begin{tabu}{|c|}
\hline S\\ \hline \langle a \rangle \\ \langle b \rangle \\ \langle c \rangle\\ \langle d \rangle \\ \hline
\end{tabu}$
\end{center}

The conjunctive query $\mc{Q} :\exists x \exists y ( S(x) \land R(x, y) \land S(y))$ posed to $D$ is true because join can be satisfied with the database tuples in different ways. For example, the tuples $S(c), R(c,b), S(b)$  jointly satisfy the query condition.
\end{multicols}
We want to identify tuples as actual causes for the query to be true. \
If the tuple $S(b)$ is deleted, the particular instatiation of the join just mentioned above becomes false. However, the query is still true, jointly via the tuples $S(a), R(a,d),$ $ S(d)$; or  the tuples $S(b), R(b,a),$ $ S(a)$. In order to falsify the query, some of these tuples have to be deleted as well. There are different combinations, but a minimum deletion is that of $S(a)$.   In more technical terms, in order for $S(b)$ to be an {\em actual cause} for the query to be true, it requires a {\em contingency set} of tuples to be further deleted. In this case,  $\{S(a)\}$ is a minimum-size contingency set, of cardinality $1$. Accordingly, $S(b)$ is an actual cause with {\em causal responsibility} $\nit{Resp}(S(b)) = \frac{1}{1 + 1} = \frac{1}{2}$.  Notice that if we just delete $S(a)$, the query is still true. \ A condition on contingency sets for an actual cause is that its deletion alone does not falsify the query; it has to be combined with the cause candidate at hand.

If we, instead, had the following database \ $D^\prime$ \ here below, \ tuple \ $S(b)$  \ would be a

\begin{multicols}{2}
\begin{center}
$\begin{tabu}{|c|}
\hline R\\ \hline \langle c, b \rangle \\ \langle a, d \rangle \\ \langle b, b\rangle \\
\hline
\end{tabu} \hspace{5mm}\begin{tabu}{|c|}
\hline S\\ \hline \langle a \rangle \\ \langle b \rangle \\ \langle c \rangle\\  \hline
\end{tabu}$
\end{center}

\noindent  {\em counterfactual cause} in the sense that it does not require any additional contingent deletion to falsify the query. The empty set is a minimum-size contingency set for it,
\end{multicols}
\noindent  and its responsibility becomes $1$, the maximum responsibility.

Tuple $S(c)$ has $\{R(b,b)\}$ as a minimum contingency set, but not $\{S(b)\}$ that alone can falsify the query. We have $\nit{Resp}(S(c)) = \frac{1}{2}$.
\boxtheorem
\end{example}

The examples above shows similarities between CBD and actual causality, in that in both cases we try to make changes with the purpose of possibly seeing a different outcome. \ In fact, a precise connection for the particular case of databases  was established  in \cite{tocs}, and the connection was appropriately exploited. We give a simple example for the gist.

\begin{example}\label{ex:mbdaex6} (example \ref{ex:mbdaex5} cont.) Instead of performing interventions directly on the data-base, and  in relation to the query at hand, we can build a CBD problem.

The first observation is that the conjunctive query $\mc{Q}\!: \ \exists x \exists y (S(x) \wedge R(x, y)  \wedge \ S(y))$ can be transformed into an {\em integrity constraint}, \ $\kappa\!: \ \neg \exists x \exists y (S(x) \wedge R(x, y)  \wedge \ S(y))$, actually a {\em denial constraint} that prohibits the satisfaction of the query join (the conjunction). \ It holds that $\mc{Q}$ is satisfied by $D$ iff $\kappa$ is violated by $D$, which would be considered the faulty behavior. \ Accordingly, we specify  a system that may not be working as normal, in the sense that the IC can be violated (under abnormal conditions). \ The main component of the diagnosis model  is the formula:
\begin{equation}
\forall x \forall y(\neg\nit{Ab}_S(x)\wedge \neg\nit{Ab}_R(x,y) \wedge \neg \nit{Ab}_S(y) \ \longrightarrow \ \neg (S(x)  \wedge R(x, y)  \wedge \ S(y)). \label{eq:kappa}
\end{equation}
Here, we have introduced one auxiliary abnormality predicate for each database predicate. This formula says that when the tuples are not abnormal, they do not participate in the violation of the IC $\kappa$. In other words,
under the normality assumptions, the database behaves as intended (or normal), i.e.  satisfying the IC, which in this case, amounts to not making the query true.

For example, the tuples $S(c), R(c,b), S(b)$  in $D$ jointly satisfy the non-negated join  on the RHS of (\ref{eq:kappa}). \ In order to make (\ref{eq:kappa}) true, at least one of  $\nit{Ab}_S(c)$, $\nit{Ab}_R(c,b)$, $ \nit{Ab}_R(b)$ has to be true. The rest continues as with Example \ref{ex:circ}.
\ The tuples whose associated abnormality atoms become true are, by definition, the actual causes for the query to be true. \boxtheorem
\end{example}

  In the database context, actual causality was also connected with abductive diagnosis for Datalog queries \cite{flairsExt}.

\section{Attribution Scores in Machine Learning}\label{sec:xai}

In machine learning, in situations such as that
 in Example \ref{ex:one}, actual causality and responsibility   have been applied to provide {\em counterfactual explanations} for classification results, and scores for them. In order to do this, having access to the internal components of the classifier is not needed, but only its input/output relation.

\begin{example} \label{ex:dos} (example \ref{ex:one} cont.) \ The entity
\begin{equation}
\e \ = \ \langle \msf{john}, \msf{18}, \msf{plumber}, \msf{70K}, \msf{harlem}, \msf{10K},\msf{basic}\rangle \label{eq:ent}
\end{equation} received  the label \ $1$ from the classifier, indicating that the loan is not granted.\footnote{We are assuming that classifiers are {\em binary}, i.e. they return labels $0$ or $1$. For simplicity and uniformity, but without loss of generality, we will assume that label $1$ is the one we want to explain.} \ In order to identify counterfactual explanations, we intervene the feature values replacing them by alternative ones from the feature domains, e.g.
\ $\e_1 \ = \  \langle \msf{john}, \ul{\msf{25}}, \msf{plumber}, \msf{70K},$ $\msf{harlem},\msf{10K}, \msf{basic}\rangle$, which receives the label  \ $0$.
\ The counterfactual version $\e_2 = \ \langle \msf{john}, \msf{18},$ $ \msf{plumber}, \ul{\msf{80K}}, \ul{\msf{brooklyn}}, \msf{10K}, \msf{basic}\rangle$ also get label \ $0$. Assuming, in the latter case, that none of the single changes alone switch the label, we could say that $\msf{Age} = \msf{25}$, so as $\msf{Income} = \msf{70K}$ with contingency $\msf{Location} = \msf{harlem}$ (and the other way around) in $\e$ are (minimal) counterfactual explanations, by being actual causes for the observed label.

We could go one step beyond, and define responsibility scores: \ $\nit{Resp}(\e,\msf{Age}) := 1$, and $\nit{Resp}(\e,\msf{Income}) := \frac{1}{2}$ (due to the additional, required, contingent change). This choice does reflect the causal strengths of attribute values in $\e$. However, it could be the case that only by changing the value of $\msf{Age}$ to $\msf{25}$ we manage to switch the label, whereas for all the other possible values for $\msf{Age}$ (while nothing else changes), the label is always {\em No}. It seems more reasonable to redefine responsibility by considering an average involving all the possible labels obtained in this way. \boxtheorem
\end{example}

The direct application of the responsibility score, as in \cite{suciu10,tocs}, works fine for explanation scores when features are  {\em binary}, say taking values $0$ or $1$, \cite{TPLP22,ijclr21}. However, when features have more than two values, it makes sense to extend the definition of the responsibility score.

\subsection{The generalized $\mathbf{\nit{Resp}}$  score}

In \cite{deem}, a generalized $\nit{Resp}$ score was introduced and investigated. We describe it next in intuitive terms, and appealing to Example \ref{ex:dos}.

\begin{enumerate}
\item For an entity $\e$ classified with  label $L(\e) =1$, \ and a feature $F^\star$, whose value $F^\star(\e)$ appears in $\e$, we want a numerical responsibility score $\nit{Resp}(\e,F^\star)$, characterizing the causal strength of $F^\star(\e)$ for outcome $L(\e)$. \ In the example, $F^\star = \msf{Salary}$, $F^\star(\e) = \msf{70K}$, and $L(\e) = 1$.

\vspace{1mm}
\item  While we keep the original value for $\msf{Salary}$ fixed, we start by defining a ``local" score for a fixed contingent assignment $\Gamma := \bar{w}$, with $F^\star \notin \Gamma \subseteq \mc{F}$. \ We define $\e^{\Gamma,\bar{w}} := \e[\Gamma:=\bar{w}]$, the entity obtained from $\e$ by changing (or redefining) its values for features in $\Gamma$, according to $\bar{w}$.

  \vspace{1mm}
    In the example, it could be $\Gamma = \{\msf{Location}\}$, and $\bar{w} := \langle \msf{brooklin}\rangle$, a contingent (new) value for $\msf{Location}$. Then, $\e^{\{\msf{Location}\},\langle \msf{brooklin}\rangle} = \e[\msf{Location}:= \msf{brooklin}]  =  \langle \msf{john}, \msf{25}, \msf{plumber}, \msf{70K},$ $\ul{\msf{brooklin}},\msf{10K},$ $ \msf{basic}\rangle$.

    \vspace{1mm}
    We make sure (or assume in the following) that $L(\e^{\Gamma,\bar{w}}) = L(\e) = 1$ holds. \ This is because, being these changes only contingent, we do not expect them to switch the label by themselves, but only and until the ``main" counterfactual change on $F^\star$ is made.

\vspace{1mm}
      In the example, we assume $L(\e[\msf{Location}:= \msf{brooklin}])  =   1$. \ Another case could be $\e^{\Gamma^{\prime}\!,\bar{w}^\prime}$\!\!, with $\Gamma^\prime = \{\msf{Activity}, \msf{Education}\}$, and $\bar{w}^\prime = \langle \msf{accountant}, \msf{medium}\rangle$, with $L(\e^{\Gamma^{\prime}\!,\bar{w}^\prime}) = 1$.

\vspace{1mm}
\item Now, for each of those $\e^{\Gamma,\bar{w}}$ as in the previous item, we consider all the different possible values $v$ for $F^\star$, while the values for all the other features are fixed as in $\e^{\Gamma,\bar{w}}$.

    For example, starting from  $\e[\msf{Location}:= \msf{brooklin}]$, we can consider $\e_1^{\prime} :=$ \linebreak $ \e[\msf{Location}:= \msf{brooklin};\ul{\msf{Salary} :=\msf{60K}}]$ \ (which is the same as $\e^{\msf{Location},\langle\msf{brooklin}\rangle}[$ $\msf{Salary}$ $ :=\msf{60K}]$), obtaining, e.g. $L(\e_1^{\prime}) = 1$. \ However, for $\e_2^{\prime} :=$ $ \e[\msf{Location}:= \msf{brooklin};$  $\ul{\msf{Salary :=80}}]$, we now obtain, e.g. $L(\e_2^{\prime}) = 0$, etc.

\vspace{1mm}
For a fixed (potentially) contingent change $(\Gamma, \bar{w})$ on $\e$, we consider the  difference between the original label $1$ and the expected label obtained by further modifying the value of $F^\star$ (in all possible ways). \ This gives us a {\em local} responsibility score, \  local to $(\Gamma, \bar{w})$: \vspace{-3mm}
\begin{eqnarray}
\nit{Resp}(\e,F^\star,\ul{\Gamma,\bar{w}}) &:=& \frac{L(\e) - \mathbb{E}(~L(\e^\prime)~|~F(\e^\prime) = \ F(\e^{\Gamma,\bar{w}}), \ \forall  F \in (\mc{F}\smallsetminus \{F^\star\}~)}{1 + |\Gamma|} \nonumber\\
&=&\frac{1 - \mathbb{E}(~L(\e^{\Gamma,\bar{w}}[F^\star := v])~|~ v \in \nit{Dom}(F^\star)~)}{1 + |\Gamma|}. \label{eq:star}
\end{eqnarray}

This local score takes into account, so as the original responsibility score in Section \ref{sec:causes}, the size of the contingency set $\Gamma$.

\vspace{1mm}
We are assuming here that there is a probability distribution over the entity population $\mc{E}$. It could be known from the start, or it could be an empirical distribution obtained from a sample.
As discussed in \cite{deem}, the choice (or whichever distribution that is available) is relevant for the computation of the general $\nit{Resp}$ score, which  involves the local ones (coming right here below).

\vspace{1mm}
\item  Now, generalizing the terms introduced in Section \ref{sec:causes},  we can say that the value $F^\star(\e)$ is
an {\em actual cause} for label $1$ when, for some $(\Gamma, \bar{w})$, (\ref{eq:star}) is positive: at least one change of value for $F^\star$ in $\e$ changes the label (with the company of $(\Gamma, \bar{w})$).

\vspace{1mm}
When $\Gamma = \emptyset$ (and then, $\bar{w}$ is an empty assignment), and (\ref{eq:star}) is positive, it means that at least one change of value for $F^\star$ in $\e$ switches the label by itself. As before, we can say that $F^\star(\e)$ is a {\em counterfactual cause}. However, as desired and expected, it is not necessarily the case anymore that counterfactual causes (as original values in $\e$) have all the same causal strength: $F_i(\e), F_j(\e)$ could be both counterfactual causes, but with different values for (\ref{eq:star}), for example if changes on the first switch the label ``fewer times" than those on the second.

\vspace{1mm}
\item Now, we can define the global score, by considering the ``best" contingencies $(\Gamma,\bar{w})$, which involves requesting from $\Gamma$ to be of minimum size: 
\begin{equation}
\nit{Resp}(\e,F^\star) \ \ := \ \max \limits_{{\Gamma, \bar{w}: \ |\Gamma| \ \\ \mbox{\scriptsize \ is min. \& } \mbox{\scriptsize (\ref{eq:star}) $> 0$}}}
\ \nit{ Resp}(\e,F^\star,\Gamma,\bar{w}). \label{eq:min}
\end{equation}
This means that we first find the minimum-size contingency sets $\Gamma$'s for which, for an associated  set of value updates $\bar{w}$, (\ref{eq:star}) becomes greater that $0$. After that, we find the maximum value for (\ref{eq:star}) over all such pairs $(\Gamma,\bar{w})$. This can be done by starting with $\Gamma = \emptyset$, and iteratively increasing the cardinality of $\Gamma$ by one, until a $(\Gamma,\bar{w})$ is found that makes (\ref{eq:star}) non-zero. We stop increasing the cardinality, and we just check if there is another $(\Gamma^\prime, \bar{w}^\prime)$ that gives a greater value for (\ref{eq:star}), with $|\Gamma^\prime| = |\Gamma|$. \ 
By taking the maximum of the local scores, we  have an existential quantification in mind: there must be a good contingency $(\Gamma, \bar{w})$, as long as $\Gamma$ has a minimum cardinality.
\end{enumerate}

With the generalized score, the difference between counterfactual and actual causes is not as relevant as before. In the end, and as discussed under Item 4. above, what matters is the size of the score. Accordingly, we can talk only about ``counterfactual explanations with responsibility score $r$". In Example \ref{ex:dos}, we could say ``$\e_2$ is a (minimal) counterfactual for $\e$ (implicitly saying that it switches the label), and the value $\msf{60K}$ for $\msf{Salary}$ is a counterfactual explanation with responsibility $\nit{Resp}(\e,\msf{Salary})$". Here, $\e_2$ is possibly only one of those counterfactual entities that contribute to making the value for $\msf{Salary}$ a counterfactual explanation, and to its (generalized) \Resp \ score.

The generalized $\nit{Resp}$ score was applied for different financial data \cite{deem}, and experimentally compared with a simpler version of responsibility, and with the $\nit{Shap}$ score \cite{lund17}, all of which can be applied with a black-box classifier, using only the input/output relation. It was also experimentally compared, with the same data, with a the FICO-score \cite{rudin} that is defined for and applied to an open-box model, and  computes scores by taking into account components of the model, in this case coefficients of nested logistic regressions.


The computation cost of the \Resp \ score is bound to be high in general since, in essence,  it explicitly involves in (\ref{eq:star}) all possible subsets of the set of features; and in (\ref{eq:min}), also the minimality condition which compares different subsets. Actually, for binary classifiers and in its simple, binary formulation, \Resp \ is already intractable \cite{TPLP22}. \ In \cite{deem}, in addition to experimental results, there is a technical discussion on the importance of the underlying distribution on the population, and on the need to perform optimized computations and approximations.

\subsection{The $\mathbf{\nit{Shap}}$ score}

The \Shap \ score was introduced in explainable ML in \cite{lund17}, as an application of the general {\em Shapley value} of {\em coalition game theory} \cite{Shapley}, which we briefly describe next.

Consider a set of players $\mc{S}$,  and a
{\em wealth-distribution  function} (or {\em game function}), $\mc{G}\!: \mc{P}(\mc{S})  \rightarrow  \mathbb{R}$, that maps subsets of $\mc{S}$ to real numbers.  \ The Shapley value of player $p \in \mc{S}$ quantifies the contribution of  $p$ to the game, for which all different coalitions are considered; each time, with $p$ and without $p$:
  \begin{equation}\nit{Shapley}(\mc{S},\mc{G},p):= \sum_{S\subseteq
  \mc{S} \setminus \{p\}} \frac{|S|! (|\mc{S}|-|S|-1)!}{|\mc{S}|!}
(\mc{G}(S\cup \{p\})-\mc{G}(S)). \label{eq:shapley}
\end{equation}
Here, $|S|! (|D|-|S|-1)!$ is the number of permutations of
$\mc{S}$ with all players  in $S$ coming first, then $p$, and then all the others. \ In other words,
this is the {\em expected contribution} of $p$ under all possible additions of $p$ to a partial random sequence of players, followed by random sequences of the rest of the players.

The Shapley value emerges as the only quantitative measure that has some specified properties in relation to coalition games \cite{roth}. It has been applied in many disciplines. \ For each particular application, one has to define a particular and appropriate game function $\mc{G}$. Close to home, it has been applied to assign scores to logical formulas to quantify their contribution to the inconsistency of a knowledge base \cite{hunter}, to quantify contributions to the inconsistency of a database \cite{benny2}, and to quantify the contribution of database tuples to making a query true \cite{shapleyDBSigmod,benny}.

In different application and with different game functions, the Shapley  value turns out to be computationally intractable,  more precisely, its time complexity is {\em $\#P$-hard} in the size of the input, c.f., for example, \cite{shapleyDBSigmod}. Intuitively, this means that it is at least as hard as any of the problems in the class $\#P$ of problems about counting the solutions to decisions problems (in $\nit{NP}$) that ask about the existence of a certain solution \cite{valiant,papadimitriou}. For example, $\nit{SAT}$ is the decision problem asking, for a propositional formula, if there exists a truth assignment (a solution) that makes the formula true. Then, $\#\!\nit{SAT}$ is the computational problem of counting the number of satisfying assignments of a  propositional formula. Clearly, $\#\!\nit{SAT}$ is at least as hard as $\nit{SAT}$ (it is good enough to count the number of solutions to know if the formula is satisfiable), and $\nit{SAT}$ is the prototypical $\nit{NP}$-complete problem, and furthermore, $\#\!\nit{SAT}$ is $\#P$-hard, actually, $\#\!P$-complete since it belongs to $\#P$.\footnote{Another $\#P$-complete problem is $\#\nit{Hamiltonian}$, about counting the number of Hamiltonian cycles in a graph. Its decision version, about the existence of a Hamiltonian cycle, is $\nit{NP}$-complete. } As a consequence, computing the Shapley value can be at least as hard as computing the number of solutions for $\nit{SAT}$; a clear indication of its high computational complexity.

As mentioned earlier in this section, the \Shap \ score is a particular case of the Shapley value in (\ref{eq:shapley}). In this case, the players are the features $F$ in $\mc{F}$, or, more precisely, the values $F(\e)$ they take for a particular entity $\e$, for which we have a binary classification label, $L(\e)$, we want to explain. The explanation comes in the form of a numerical score for $F(\e)$, reflecting its relevance for the observed label. Since all the feature values contribute to the resulting label, we may conceive the features values as players in a coalition game.

 The game function, for a given subset $S$ of the features, is the {\em expected (value of the) label} over all possible entities whose values coincide with those of $\e$ for the features in $S$: \
\begin{equation}
\mc{G}_\mathbf{e}(S) := \mathbb{E}(L(\mathbf{e'})~|~\e^\prime \in \mc{E} \ \mbox{ and } \ \e^{\prime}_{\!S} = \e_S),
\end{equation} where  $\e^\prime_S, \e_S$ denote the projections of $\e^\prime$ and $\e$ on $S$, resulting in two subrecords of feature values. \ We can see that the game function depends on the entity at hand $\e$.

With the game function in (\ref{eq:shapley}), we obtain the \Shap \ score for a feature value $F^\star(\e)$ in $\e$:
\begin{eqnarray}\nit{Shap}(\mc{F},\mc{G}_\mathbf{e},F^\star) &:=&
\!\!\!\!\!\sum_{S\subseteq
  \mc{F} \setminus \{F^\star\}} \frac{|S|! (|\mc{F}|-|S|-1)!}{|\mc{F}|!}
[ \mathbb{E}(L(\e^\prime|\e^{\prime}_{S\cup \{F^\star\}} = \e_{S\cup \{F^\star\}}) - \nonumber \\
&&\hspace*{4.25cm}\mathbb{E}(L(\e^\prime)|\e^{\prime}_S = \e_S)]. \label{eq:shap}
\end{eqnarray}

\begin{example} \label{ex:tres} (example \ref{ex:dos} cont.) \
For the fixed entity $\e$ in (\ref{eq:ent}) and feature $F^\star = \msf{Salary}$, one of the terms in (\ref{eq:shap}) is obtained by considering $S = \{\msf{Location}\} \subseteq \mc{F}$:
\begin{eqnarray*}
&\frac{|1|! (7-1-1)!}{7!} \times
(\mc{G}_\e(\{\msf{Location}\}\cup \{\msf{Salary}\})-\mc{G}_\e(\{\msf{Location}\}))&\\
&\hspace*{3.5cm}= \ \frac{1}{42} \times (\mc{G}_\e(\{\msf{Location}, \msf{Salary}\})-\mc{G}_\e(\{\msf{Location}\})),&
\end{eqnarray*}
with, e.g., \ $\mc{G}_\e(\{\msf{Location}, \msf{Salary}\})= \mbb{E}(L(\e^\prime)~|~ \e^\prime \in \mc{E}, \  \msf{Location}(\e^\prime) = \msf{harlem}, \mbox{ and }$ $ \msf{Salary}(\e^\prime) = \msf{70K})$, \  that is, the expected label over all entities that have the same  values as $\e$ for features $\msf{Salary}$ and $\msf{Location}$. \ Then, \
$\mc{G}_\e(\{\msf{Location}, \msf{Salary}\})-\mc{G}_\e(\{\msf{Location}\})$ is the expected  difference in the label between the case where the values for $\msf{Location}$ and $\msf{Salary}$ are fixed as for $\e$, and the case where only the value for $\msf{Location}$ is fixed as in $\e$, measuring a local contribution of $\e$'s value for $\msf{Salary}$.  After that, all these local differences are averaged over all subsets $S$ of $\mc{F}$, and the permutations in which they participate. \boxtheorem
\end{example}

We can see that, so as the \Resp \ score, \Shap \ is a {\em local} explanation score, for a particular entity at hand $\e$.  \ Since the introduction of \Shap \ in this form, some variations have been proposed. So as for \Resp,  \Shap \ depends, via the game function, on an underlying probability distribution on the entity population $\mc{E}$. The distribution may impact not only the \Shap \ scores, but also their computation \cite{deem}.

\subsection{Computation of the $\mathbf{\nit{Shap}}$ score}

Boolean classifiers, i.e. propositional formulas with binary input features and binary labels, are particularly relevant, {\em per se} and because  they can represent other classifiers by means of appropriate encodings. For example, the circuit in Figure \ref{fig:circ} can be seen as a binary classifier that can be represented, on the basis of (\ref{eq:cir}), by means of the single propositional formula \
$ ((a \wedge
b) \vee
c)$ \ that, depending on the binary values for $a,b,c$, also returns a binary value.

Boolean classifiers, as logical formulas, have been extensively investigated. In particular, much is known about the satisfiability problem of propositional formulas, $\nit{SAT}$,  and also about the {\em model counting} problem, i.e. that of counting the number of satisfying assignments, denoted $\#\!\nit{SAT}$.
\ In the area of {\em knowledge compilation}, the complexity of $\#\!\nit{SAT}$ and other problems in relation to the syntactic form of the Boolean formulas have been investigated  \cite{darwicheKC,darwicheJANCL,selman}. \ Boolean classifiers turn out to be quite relevant to understand and investigate the complexity of \Shap \ computation.

The computation of \Shap \ is bound to be expensive, for similar reasons as for \Resp. For the computation of both, all we need is the input/output relation of the classifier, to compute labels for different alternative entities (counterfactuals). However, in principle, far too many combinations have to go through the classifier. Actually, under the {\em product probability distribution} on $\mc{E}$ (which assigns independent probabilities to the feature values), even with an explicit, open classifier for binary entities, the computation of \Shap \ can be intractable.

In fact, as shown in \cite{deem}, for Boolean classifiers in the class $\nit{Monotone}2\nit{CNF}$, of negation-free propositional formulas in conjunctive normal form with at most two atoms per clause, \Shap \ can be $\#\!P$-hard. This is obtained via a polynomial reduction from $\#\nit{Monotone}2\nit{CNF}$, the problem of counting the number of satisfying assignments for a formula in the class, which is known to be $\#\!P$-complete \cite{valiant}.\footnote{Interestingly, the decision version of the problem, i.e. of deciding if a formula in $\nit{Monotone}2\nit{CNF}$ is satisfiable, is trivially tractable: the assignment that makes all atoms true satisfies the formula.} \ For example, if the classifier  is \ $(x_1 \vee x_2) \wedge (x_2 \vee x_3)$, which belongs to $\#\nit{Monotone}2\nit{CNF}$, the entity $\e_1 = \langle 1,0,1\rangle$ (with values for $x_1, x_2, x_3$, in this order) gets label $1$, whereas the entity $\e_2 = \langle 1,0,0\rangle$ gets label $0$. \ The number of satisfying truth assignments, equivalently, the number of entities that get label $1$, is $5$, corresponding to $\langle 1,1,1\rangle$, $\langle 1,0,1\rangle$,  $\langle 0,1,1\rangle$,
 $\langle 0,1,0\rangle$, and $\langle 1,1,0\rangle$.

Given that \Shap \ can be $\#P$-hard, a natural question is whether for some classes of open-box classifiers one can compute \Shap \ in polynomial time in the size of the model and input. The idea is to try to take advantage of the internal stricture and components of the classifier -as opposed to only the input/output relation of the classifier- in order to compute \Shap \ efficiently. We recall from results mentioned earlier in this section that having an open-box model does not guarantee tractability of \Shap. Natural classifiers that have been considered in relation to a possible tractable computation of \Shap \ are decision trees and random forests \cite{lundberg20}.

The problem of tractability of \Shap \ was investigated in detail in \cite{AAAI21}, and through other methods also in \cite{guyAAAI21}. They briefly describe the former approach in the rest of this section.
Tractable and intractable cases were identified, with algorithms for the tractable cases. (Approximations for the intractable cases were further investigated in \cite{AAAI21ext}.) In particular, the tractability for decision trees and random forests was established, which required first identifying the right abstraction that allows for a general proof, leaves aside contingent details, and is also broad enough to include interesting classes of classifiers.

In \cite{AAAI21}, it was proved that, for a Boolean classifier $L$ (identified with its label, the output gate or variable), the uniform distribution on $\mc{E}$, and $\mc{F} = \{F_1, \ldots, F_n\}$:
\begin{equation}\#\!\nit{SAT}(L) \ = \ 2^{|\mc{F}|} \times (L(\e) - \sum_{i=1}^n \nit{Shap}(\mc{F},G_{\e},F_i)). \label{eq:red}
\end{equation}

\vspace{-3mm}
This result makes, under the usual complexity-theoretic assumptions, impossible for \Shap \ to be tractable for any circuit $L$ for which $\#\!\nit{SAT}$ is intractable. (If we could compute \Shap \ fast, we could also compute $\#\!\nit{SAT}$ fast, assuming we have an efficient classifier.) \ This excludes, as seen earlier in this section, classifiers that are in the class $\nit{Monotone}2\nit{CNF}$. Accordingly, only classifiers in a more amenable class  became candidates, with the restriction that the class should be able to accommodate interesting classifiers. That is how the class of  {\em deterministic and decomposable Boolean circuits} (dDBCs) became the target of investigation.

\begin{figure}
\centerline{\includegraphics[width=5cm]{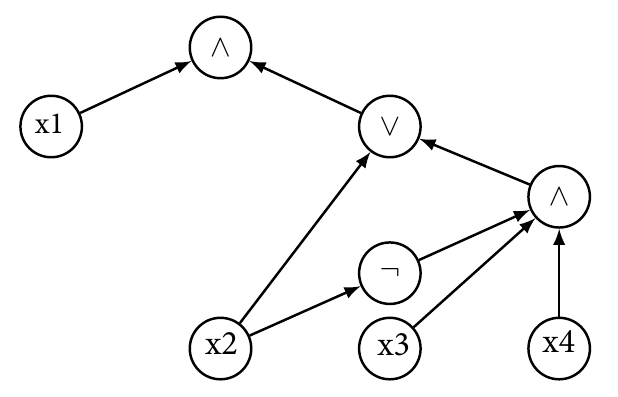}}\vspace{-2mm}
\caption{A decomposable and deterministic Boolean classifier}\label{fig:dDBC}
\end{figure}

Each $\vee$-gate of a dDBC can have only one of the disjuncts true (determinism), and for each $\wedge$-gate,
the conjuncts do not share variables (decomposition). Nodes are labeled with $\vee, \wedge$, or $\neg$ gates, and input gates with features (propositional variables) or binary constants. An example of such a classifier, borrowed from \cite{AAAI21},  is shown in Figure   \ref{fig:dDBC}, which has four (input) features, and an a gate that returns the output label (the $\wedge$ at the top). \ For a counterexample, the BC $((a \wedge b) \vee c)$ for Figure \ref{fig:circ} is decomposable, but not deterministic.\footnote{It could be transformed into a dDBC, but this would make the circuit grow. The transformation cost  is always a concern in the area of knowledge compilation. For some classes of BCs, a transformation into another class could take exponential time; sometimes exponential on a fixed parameter, etc. \cite{vardi,monet}. }

Model counting is known to be tractable for dDBCs. However, this does not imply (via (\ref{eq:red}) or any other obvious way) that \Shap \ is tractable. It is also the case that relaxing any of the determinism or decomposibility conditions makes model counting $\#P$-hard  \cite{AAAI21ext}, preventing \Shap \ from being tractable.

It turns out that $\nit{Shap}$ computation is tractable for dDBCs (under the uniform or the product distribution), from which we also get tractability of $\nit{Shap}$ for free for a vast collection of classifiers that can be efficiently compiled into (or represented as) dDBCs; among them we find:  \ Decision Trees  (even with non-binary features), Random Forests, Ordered Binary Decision Diagrams (OBDDs) \cite{bryant}, Deterministic-Decomposable Negation Normal-Forms  (dDNNFs), Binary Neural Networks (e.g. via OBDDs) \cite{darwicheKR20}, etc.

\begin{figure}
\centerline{\includegraphics[width=4cm]{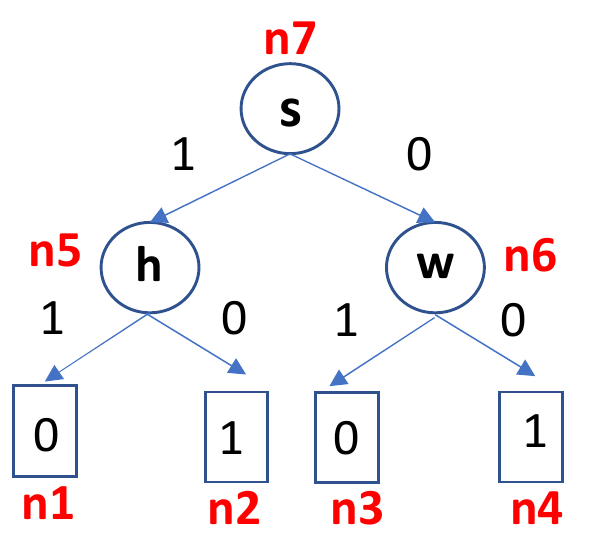}~~~~~~~~~~~~~~~\includegraphics[width=4.5cm]{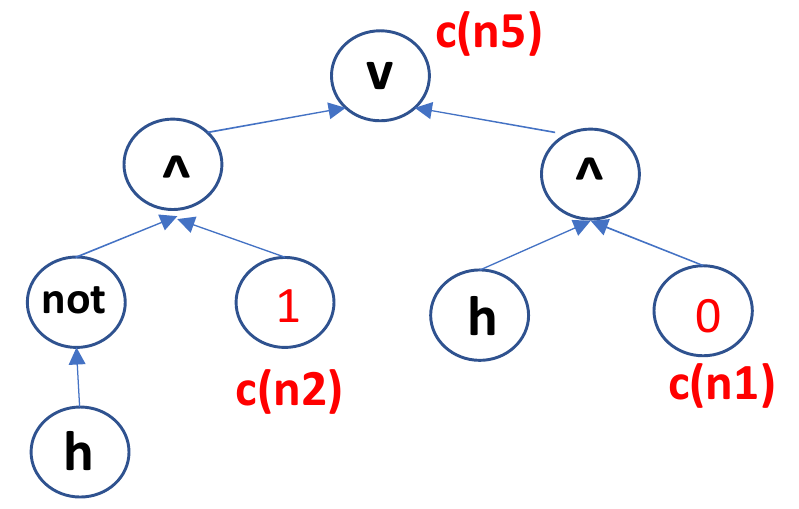}}
\vspace{-2mm}\caption{A decision tree and part of its compilation into an dDBC}\label{fig:dt}
\end{figure}

For the gist, consider the binary decision tree (DT) on the LHS of Figure \ref{fig:dt}. It can be inductively and efficiently compiled into a dDBC \cite[appendix A]{AAAI21ext}. The leaves of the DT
become labeled with propositional constants, $0$ or $1$. Each node, $n$, is compiled into a circuit $c(n)$, and the final dDBC corresponds to the compilation, $c(r)$, of the root node $r$, in this case, $c(n7)$ for node $c7$. \ Figure \ref{fig:dt} shows on the RHS, the compilation $c(n5)$ of node $n5$ of the DT. \ If the decision tree is not binary, it is first binarized, and then compiled \cite[sec. 7]{AAAI21ext}.

Abductive explanations have for  dDNNF-Boolean circuits have been investigated in \cite{marques-SilvaAAAI22}.

 \section{Counterfactual Reasoning}

As we saw in the preceding sections,  counterfactuals are at the basis of the responsibility score. It is captured in (\ref{eq:star}) that some counterfactual interventions may not change the label of a feature value. However, for a feature value to have a non-zero responsibility, there must be at least one counterfactual
intervention on that value that changes the label. In the case of \Shap, all counterfactuals are implicit and taken into account in its computation.

Independently from their participation in the definition and computation of attribution scores, counterfactuals are relevant and informative {\em per se} \cite{verma}. Assuming they satisfy some minimality condition, they tell us what change of feature values may change the label, pointing to the relevance of the original values for the label at hand. Furthermore, in many situations we would like to know what we could do in order to change the label, e.g. granting a loan instead of rejecting it. If those changes can be usefully made in practice, we are talking about {\em actionable counterfactuals}, or counterfactuals that are {\em resources} \cite{ustun,valera,karimi,verma}. In the ongoing example, changing the age, and then waiting for seven years to get the loan, may not be feasible. Nor changing the name of the applicant from $\msf{john}$ to $\msf{elon}$. However, slightly increasing the salary might be doable. \  We may also want to compare two alternative counterfactuals, or ask about the existence of one with a particular property, e.g. a particular value for a feature, etc.

Working with and analyzing counterfactuals can be made easier and more understandable if we can {\em reason} about them under a single platform that includes (or interacts with) the classifier, and the score computation mechanisms, for a particular application.  By reasoning we mean, among other tasks, the {\em logical specification} of counterfactuals, the {\em entailment} of counterfactuals, their analysis, and obtaining them and their properties. Counterfactual reasoning has been investigated from a logical point of view \cite{eiterGottlob}. \ If we want to put counterfactual reasoning in practice, we need the right logics and their implementations.

We have recently argued in favor of using {\em answer-set programming} (ASP) for this task \cite{RW21}. ASP is a form of  logic programming that has several advantages for the kind of problems we are confronting \cite{brewka}, among them: (a) It has the right-expressive power and computational complexity since it can be applied to solve complex combinatorial problems. (b) Through its non-monotonic negation, and associated predicate minimality, it allows to represent the (common sense) inertia or persistence of objects and their properties unless they are explicitly changed (intervened in our case). (c) Logical disjunction allows to specify several alternative counterfactual candidates at once. (d) One can pose queries to obtain results from reasoning (c.f. below). (e) Its {\em possible worlds semantics} for the specification leads to multiple models corresponding to equally multiple counterfactuals, with different properties. (f) A {\em cautious} and a {\em brave query answering semantics} that allow asking what is true {\em in all} or {\em in some} models, respectively. (g) There are implementations of ASP that can interact with external classifiers.

In essence, one can specify the underlying logical setting, e.g. some classes of classifiers, the possible interventions that lead to counterfactuals, properties such as actionability and other properties that may force counterfactuals ``to make sense", minimality conditions on counterfactuals, and the computation of (some) attribution-scores that are based on counterfactuals. ASPs also allow for score aggregations, for more global and higher-level explanations.

 We have used ASP, in particular the {\em DLV} system and its various extensions \cite{dlv}, to specify counterfactuals for causality in data management \cite{kais} and for explanations in ML \cite{ijclr21,TPLP22}, including the computation of  the simple responsibility score.\footnote{It is worth mentioning that ASP and {\em DLV} have been used to specify and compute model-based diagnoses, both in their abductive and consistency-based formulations \cite{eiterDiag}.} \ However, if we want to compute scores that are defined in terms of of expected values, e.g. \Resp \ or \Shap, over probability distributions  other than easily specifiable and computable ones, it may be necessary to resort to probabilistic extensions of ASP \cite{pASP,lee,lee22,plingo}.

Counterfactual reasoning on the basis of ASPs is best seen as {\em query-driven}. One can pose all kinds of queries about counterfactuals, for which having a cautious and a brave semantics comes handy. Typical queries request counterfactual with a particular property, e.g. that changes, or does not change a particular feature value; or counterfactuals that exhibit (or avoid) a particular combination of  feature values; or pairs of counterfactuals that differ in a pre-specified manner. Queries may ask if a particular feature value is never changed in a (preferred kind of) counterfactual; or about the existence of good counterfactuals that do not change more than a certain number of feature values; or whether there are ``similar" counterfactuals with different labels (according to a specified notion of similarity), etc. \ C.f. \cite{RW21,ijclr21,TPLP22} for concrete examples. One can also compare counterfactual entities with pre-specified {\em reference entities} for obtaining {\em contrastive explanations} \cite{karimi,miller}.

Specifying actionable counterfactuals is only one way to convey relevant {\em application domain knowledge} into the definition of counterfactuals. A logical specification allows for much more than that, including the adoption of  a {\em domain semantics}. We started this section recalling that, in principle, all counterfactuals are considered for the computation of attribution scores. However, it would be much more natural, useful, and possibly also more efficient, to consider and compute counterfactuals that conform to the domain semantics, which can be specified through logical rules and constraints, in such a way that only those are brought into a score computation \cite{TPLP22}. For example, since the age of an individual (represented as an entity) never decreases, changing it by a lower value may not make much sense in  most applications. Similarly, some combinations of values may not make sense, e.g. $\msf{Age} = \msf{6}$ \ and \ $\msf{MaritalStatus} = \msf{married}$.

Considering that counterfactuals are at the very basis of causality, it is not surprising to see that they are playing a prominent role in Explainable AI. However, they have also found applications  in Fairness in AI. C.f. \cite{roy} and references therein.

\vspace{2mm}\noindent {\bf Acknowledgements:} \ Part
of this work was funded by ANID - Millennium Science Initiative Program -
Code ICN17002. \ The author is a Professor Emeritus at Carleton University, Ottawa, Canada; and a Senior Universidad Adolfo Ib\'a\~nez (UAI) Fellow,  Chile. \ Comments by Paloma Bertossi on an earlier version of the article are much appreciated.

\vspace{-2mm}
\bibliographystyle{plain}

\end{document}